\documentclass[lettersize,journal]{IEEEtran}
\usepackage{amsmath,amsfonts}
\usepackage{algorithmic}
\usepackage{array}
\usepackage[caption=false,font=normalsize,labelfont=sf,textfont=sf]{subfig}
\usepackage{textcomp,color}
\usepackage{stfloats}
\usepackage{url}
\usepackage{verbatim}
\usepackage{graphicx}
\usepackage{cite}
\usepackage{amssymb}
\usepackage{bm}
\usepackage{booktabs}
\usepackage[ruled, linesnumbered]{algorithm2e}
\usepackage{makecell}
\usepackage{multirow}
\newcommand{\methodname}{{\tt{SPD-CFL}}}

\usepackage{soul}
\soulregister\cite7
\soulregister\ref7

\usepackage{color} 
\begin{document}
% ...
\makeatletter 
\let\myorg@bibitem\bibitem
\def\bibitem#1#2\par{%
  \@ifundefined{bibitem@#1}{%
    \myorg@bibitem{#1}#2\par
  }{%
    \begingroup
      \color{\csname bibitem@#1\endcsname}%
      \myorg@bibitem{#1}#2\par
    \endgroup
  }%
}

\makeatother 
% ...


\title{\methodname{}: Stepwise Parameter Dropout for Efficient Continual Federated Learning}

\author{Yuning Yang, Han Yu, \IEEEmembership{Senior Member, IEEE}, Chuan Sun, Tianrun Gao, Xiaohong Liu, \\Xiaodong Xu, \IEEEmembership{Senior Member, IEEE}, Ping Zhang, \IEEEmembership{Fellow, IEEE}, and Guangyu Wang

\thanks{This research is supported, in part, by the National Natural Science Foundation of China (grant 62272055); New Cornerstone Science Foundation through the XPLORER PRIZE; Young Elite Scientists Sponsorship Program by CAST (2021QNRC001); in part by the China Scholarship Council; and the National Research Foundation Singapore and DSO National Laboratories under the AI Singapore Programme (No. AISG2-RP-2020-019). (\textit{Corresponding author: Guangyu Wang})}

\IEEEcompsocitemizethanks{
    \IEEEcompsocthanksitem Yuning Yang, Tianrun Gao, and Guangyu Wang are with the State Key Laboratory of Networking and Switching Technology, Beijing University of Posts and Telecommunications, Beijing 100876, China (e-mail: \{yangyuning, tianrun.gao, guangyu.wang\}@bupt.edu.cn).
    \IEEEcompsocthanksitem Han Yu and Chuan Sun are with the College of Computing and Data Science, Nanyang Technological University, 50 Nanyang Avenue, 639798, Singapore (e-mail: \{han.yu, chuan.sun\}@ntu.edu.sg).
    \IEEEcompsocthanksitem Xiaohong Liu is with the UCL Cancer Institute, University College London, WC1E 6DD London, U.K. (e-mail: xhliu17@gmail.com).
    \IEEEcompsocthanksitem Xiaodong Xu and Ping Zhang are with the State Key Laboratory of Networking and Switching Technology, Beijing University of Posts and Telecommunications, Beijing 100876, China, and also with the Department of Broadband Communication, Peng Cheng Laboratory, Shenzhen 518066, Guangdong, China (e-mail: \{xuxiaodong, pzhang\}@bupt.edu.cn).
}
}

\markboth{Journal of \LaTeX\ Class Files,~Vol.~14, No.~8, August~2021}%
{Shell \MakeLowercase{\textit{et al.}}: A Sample Article Using IEEEtran.cls for IEEE Journals}

\IEEEpubid{0000--0000/00\$00.00~\copyright~2021 IEEE}

\maketitle

\begin{abstract}

Federated Learning (FL) is a collaborative machine learning paradigm for training models on local sensitive data with privacy protection. Pre-trained transformer-based models have emerged as useful foundation models (FMs) to be fine-tuned for a wide range of downstream tasks. However, large-scale pre-trained models make it challenging for traditional FL due to high communication overhead in the resource-constrained IoT. This has inspired the field of parameter-efficient fine-tuning (PEFT) research. Existing PEFT methods generally assume that a target parameter dropout rate is predefined before FL training. They then attempt to optimize model performance at the given dropout level. Such an approach places the burden on human users to find a dropout rate that provides a satisfactory level of performance through trial-and-error, which is time consuming and resource intensive.
To address this limitation, we propose the \underline{S}tep-wise \underline{P}arameter \underline{D}ropout for \underline{C}ontinual \underline{F}ederated \underline{L}earning (\methodname{}) approach. Instead of pre-defining a desired dropout rate, it allows users to specify the target level of performance and then attempts to find the most suitable dropout rate for the given FL model. Specifically, on the server side, \methodname{} drops trainable parameters in a stepwise manner to improve communication efficiency by reducing the rank of low-rank adaptation (LoRA). The sensitivity-based gradient consistency (SGC) measure is designed to facilitate the adaptive adjustment of parameter dropout. In addition, \methodname{} introduces continual learning (CL) on the client side to mitigate performance degradation due to the inconsistent optima with distinct parameter dropout rates under heterogeneous FL. By aligning the current local LoRA modules with the previous and current global ones, \methodname{} flexibly trades off model stability versus plasticity via the regularization strategy. Extensive experiments on the public benchmark dataset CIFAR-10 and a real-world medical Face dataset demonstrate significant superiority of \methodname{} over state-of-the-art methods. Compared to the best-performing baseline, it achieves a 2.07\% higher test AUC while reducing communication overhead by 29.53\%.

\end{abstract}

\begin{IEEEkeywords}
Federated learning, Parameter-efficient fine-tuning, Stepwise parameter dropout, Continual learning.
\end{IEEEkeywords}

\section{Introduction}

Federated learning (FL) \cite{Kairouz-et-al:2021} offers a promising privacy-preserving distributed machine learning solution to satisfy the requirements of privacy concerns and regulatory restrictions in the field of sensitive Internet of Things (IoT) like finance and medicine, enabling multiple data owners to collaboratively train models without exposing local data \cite{tangefficient, liu2023ai, guo2021byzantine}. The success of pre-trained foundation models (FMs), such as BERT and GPT, has reduced the need to train models from scratch, accelerating FL convergence and delivering superior performance for various downstream tasks \cite{qu2022rethinking, chenimportance, tian2022fedbert}. Specifically, local clients can download the global pre-trained model from the central FL server, and update the model based on their datasets. The local model parameters are then uploaded to the server to aggregate a new global model for future training and inference \cite{yangdense, gao2023fedmbp}. However, the iterative transmission of large pre-trained model parameters incurs high communication overhead, posing significant challenges for bandwidth-limited and latency-sensitive FL over IoT. This has inspired the field of communication-efficient FL, particularly parameter-efficient fine-tuning (PEFT) \cite{sun2022conquering, chen2022fedtune}.
\IEEEpubidadjcol

\begin{figure}[t]
    \centering
    \includegraphics[scale=0.54]{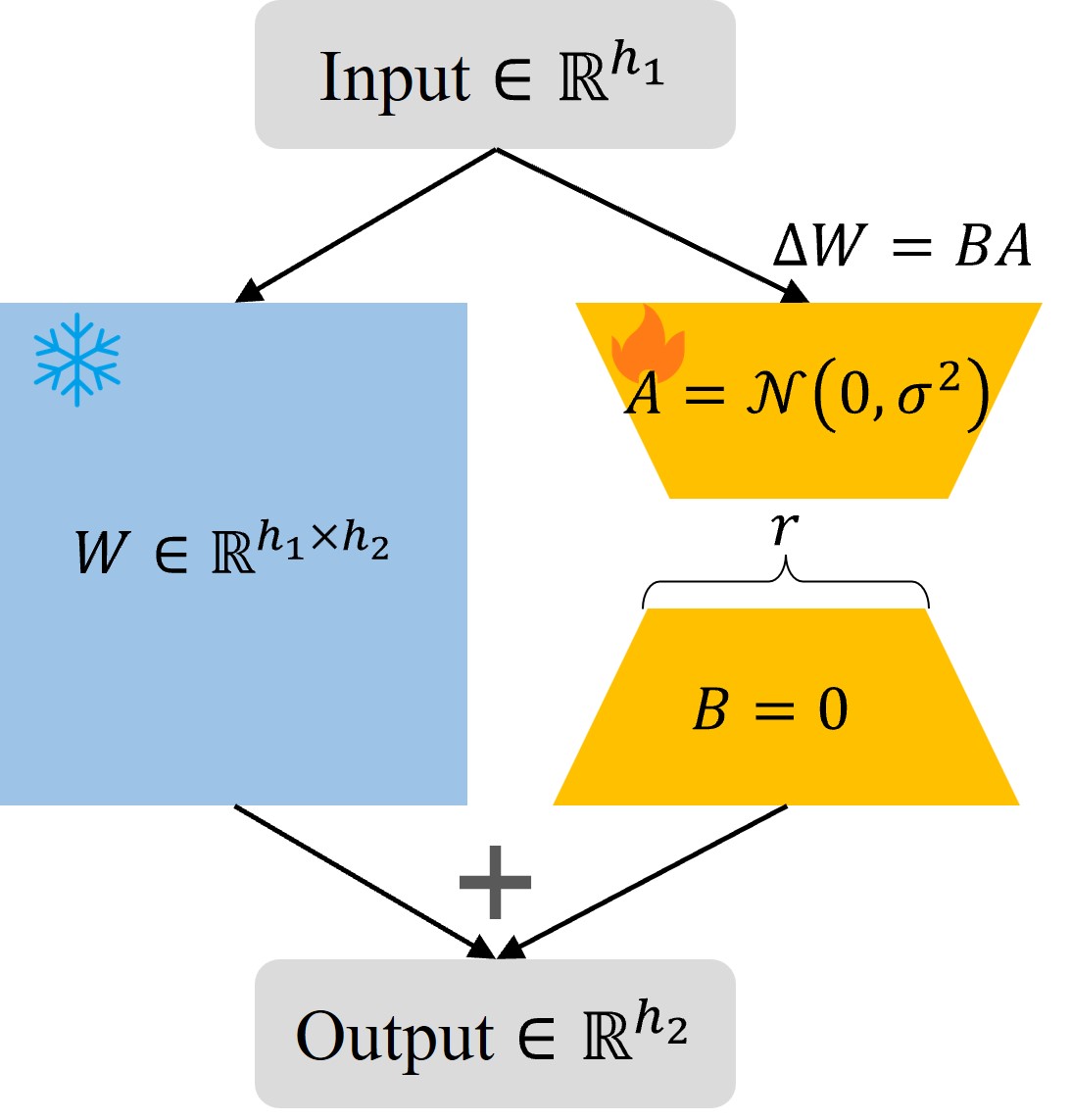}
    \caption{Structure of LoRA. The pre-trained weight matrix $W$ is frozen, while the injected low-rank decomposition matrix (LoRA module) $\Delta W=BA$ is updated during training.}
    \label{fig:LoRA}
\end{figure}

\begin{figure}[t]
    \centering
    \includegraphics[scale=0.4]{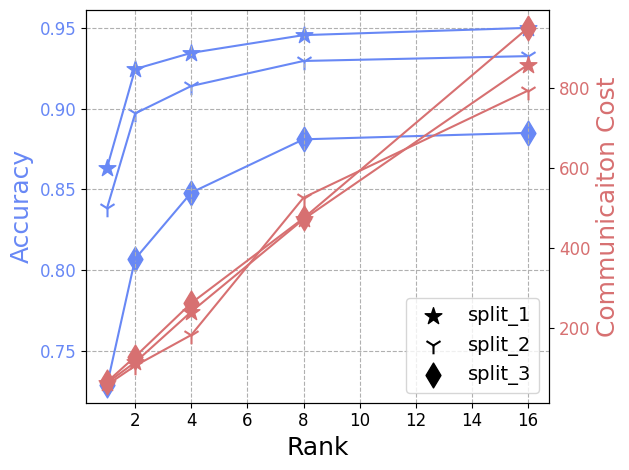}
    \caption{The accuracy and communication cost of LoRA with varying rank under different non-IID settings. 'Split\_1', 'Split\_2' and 'Split\_3' represent the IID, moderate non-IID and extreme non-IID FL, respectively.}
    \label{fig:LoRA_rank}
\end{figure}

Previous research on communication-efficient FL (e.g., model pruning, compression, parameter dropout) faces the challenges of substantial maintenance costs and poor model scalability, particularly when applied to large-scale pre-trained models \cite{shahid2021communication,almanifi2023communication}. Recently, PEFT has emerged as a promising solution due to its flexible integration and ability to reduce communication overhead by fine-tuning and sharing a small number of parameters within resource-constrained IoT systems \cite{ding2023parameter,han2024parameter,ren2024advances}. AdaLoRA\cite{zhang2023adalora} and IncreLoRA\cite{zhang2023increlora} further allocate parameter budgets adaptively by assigning more parameter budgets to important modules for learning fine-grained information while allocating fewer budgets to unimportant modules to reduce parameter overhead. Cho et al. \cite{choheterogeneous} and Bai et al. \cite{bai2024federated} introduced PEFT to practical scenarios of heterogeneous FL, aiming at balancing resource allocation and computational capacity across clients. Despite these advantages, FL with PEFT still attempts to optimize model performance guided by the given target parameter dropout rate during the training process. 
To illustrate this, we focus on Low-Rank Adaptation (LoRA), as depicted in Figure \ref{fig:LoRA}, which introduces low-rank matrices alongside the pre-trained model without incurring additional inference latency. By artificially adjusting the rank of LoRA under varying non-IID conditions, our experimental results in Figure \ref{fig:LoRA_rank} demonstrate that the lower parameter dropout rate (i.e., higher rank) with higher communication overhead can typically lead to superior performance. This phenomenon is more pronounced in non-IID settings (e.g., Split\_3). Therefore, the trial-and-error approach incurs substantial human and computational costs in searching for a parameter dropout rate that achieves the desired model performance.

To address these challenges, we propose the \underline{S}tep-wise \underline{P}arameter \underline{D}ropout for \underline{C}ontinual \underline{F}ederated \underline{L}earning (\methodname{}) approach to balance performance with communication efficiency over time. Motivated by model pruning and adaptive parameter budget allocation, \methodname{} employs gradient-based stepwise parameter dropout on the server side to enhance FL's communication efficiency. Specifically, \methodname{} initially assigns a lower parameter dropout rate to capture fine-grained and task-specific information, then progressively drops more trainable parameters to reduce communication costs by decreasing the rank of LoRA. Prior research has demonstrated that gradients across clients initially converge uniformly toward the global optimum, but tend to diverge as training progresses \cite{chen2023gift}. Building on this insight, we design a novel Sensitivity-based Gradient Consistency (SGC) metric to monitor training dynamics. An increase in SGC indicates greater training stability and triggers parameter dropout. To mitigate performance degradation caused by inconsistent optimization objectives with varying parameter dropout rates under the non-IID FL, \methodname{} incorporates continual learning (CL) via a regularization strategy in federated fine-tuning on the client side. In addition, the CL regularizer aligns the current local LoRA module with the global LoRA module with the accumulated previous parameters to reduce catastrophic forgetting (stability), while enhancing the consistency between the current local and global LoRA modules to improve optimization for new objectives (plasticity). In general, our \methodname{} offers a more effective solution for enhancing communication efficiency while maintaining model performance, enabling the search for an optimal parameter dropout rate until satisfactory performance is achieved.

We conducted extensive experiments on the public CIFAR-10 benchmarking dataset and a private medical Face dataset. The results demonstrate the significant superiority of \methodname{} over six state-of-the-art baselines. Compared to the best-performing existing approach, \methodname{} achieves a 2.07\% higher test AUC while reducing communication overhead by 29.53\%.

\section{Related Work}

\subsection{Federated PEFT}

Given its strong transferability and adaptability to downstream tasks, PEFT has emerged as a prevalent and rapidly advancing technique for optimizing large pre-trained models by training a minimal number of parameters \cite{xin2024parameter}. PEFT methods can be categorized into three main groups: reparameterization-based methods, additive methods, and selective methods \cite{lialin2023scaling}. Reparameterization-based approaches (e.g., LoRA, Factor-Tuning (FacT)) reduce the number of trainable parameters by leveraging low-rank representations \cite{hulora, jie2023fact}. Additive methods, including Adapters, introduce a small number of additional parameters between adjacent transformer layers and tune only these newly inserted modules \cite{cai2022fedadapter, guig}. In contrast, selective methods update only a small subset of critical components within the pre-trained model, such as bias terms \cite{gheini2021cross, wangoverwriting}. Among these, LoRA stands out for its advantage of introducing no inference latency and has been widely adopted for various tasks. For instance, LoRA variants have developed adaptive parameter allocation techniques, where important modules are allocated more parameter budgets to capture fine-grained knowledge, while unimportant modules are assigned fewer budgets to improve parameter efficiency \cite{zhang2023adalora, zhang2023increlora}.

While PEFT was originally motivated by the need to reduce the communication burden in FL without sacrificing performance, its application in FL systems presents unique challenges, particularly in addressing issues such as non-IID data and system heterogeneity. Early works, such as FedTune \cite{chen2022fedtune} and FedPEFT \cite{sun2022conquering}, explored the use of PEFT in FL, focusing primarily on prompts, biases, heads, and adapters, but they did not include the valuable LoRA method. Cai et al. introduced FedAdapter to examine the configuration of adapter depth and width for natural language processing (NLP) tasks in FL, which are sensitive to convergence speed and efficiency \cite{cai2022fedadapter}. Although effective, the generalizability of FedAdapter to other PEFT methods remains unclear. As PEFT continues to evolve within FL, researchers have recognized that the inherent heterogeneity of FL systems can degrade performance. In response, methods such as HETLoRA \cite{choheterogeneous} and FlexLoRA \cite{bai2024federated} have addressed system heterogeneity by allowing LoRA ranks to vary across clients based on their available resources and computational capacities. Other approaches, such as FedLoRA \cite{yi2023pfedlora} and PERAD \cite{xie2024perada}, have leveraged personalized FL to mitigate data heterogeneity by distilling generalized knowledge from personalized local adapters. To bridge the gap between full fine-tuning and PEFT in heterogeneous FL, SLoRA \cite{babakniya2023slora} introduced a sparse fine-tuning stage that applies a random data-independent binary mask with uniform density across all layers before the typical LoRA stage.

However, in the methods mentioned above, model performance is optimized for a given parameter dropout rate, making it time- and resource-consuming to find a dropout rate with satisfactory performance. Innovated by the model pruning and adaptive parameter budget allocation, we propose a stepwise parameter dropout approach for federated fine-tuning, which adaptively drops trainable parameters based on gradient dynamics.

\subsection{Continual Learning}

Catastrophic forgetting is a critical challenge in multi-task learning, where adaptation to new tasks leads to performance degradation on previous tasks due to discrepancies in their optimal objectives \cite{ren2024analyzing, bafghi2024parameter}. CL offers a promising solution for maintaining old knowledge by pushing new models towards previous optima. CL is generally categorized into three paradigms: Task-Incremental Learning (TIL), Class-Incremental Learning (CIL), and Domain-Incremental Learning (DIL) \cite{kim2023achieving, wangcomprehensive}. In TIL, task identities are provided during both training and testing, and the label spaces are disjoint across tasks \cite{gao2023enhancing}. In CIL, although label spaces remain disjoint, the model must infer the task during testing, as task identities are only available during training \cite{chengeneral}. In DIL, the label space remains constant, but input distributions differ across tasks, and task identities are not provided at any stage, requiring the model to focus on solving one task at a time \cite{lamers2023clustering}.

Regardless of the specific CL type, all approaches strive to learn from new tasks (plasticity) while retaining knowledge from earlier tasks (stability), employing methods such as parameter isolation, memory replay, and regularization-based techniques \cite{wickramasinghecontinual}. Parameter isolation methods mitigate task interference by allocating distinct sets of model parameters to different tasks \cite{tiwari2022gcr}. Memory replay methods store a subset of previous task examples in a buffer and sample from this memory when training on new tasks \cite{zhang2022simple}. Regularization-based methods, which are widely studied, control model parameters, hyperparameters, or activations to adjust the update direction for new tasks while preserving information from previous tasks through simple regularizers \cite{kirkpatrick2017overcoming, aljundi2018memory}.

Approaches such as ANCL \cite{kim2023achieving} and AFEC \cite{wang2021afec} optimize model performance by introducing two regularization terms to balance plasticity and stability. Building on this idea, our work uses regularization-based continual learning to address the inconsistencies in optimal objectives caused by parameter dropout. Specifically, we align the local LoRA modules with both the accumulated past and current global LoRA modules.

\section{The Proposed \methodname{} Approach}

\subsection{Overview}

\begin{figure}[t]
    \centering
    \includegraphics[width=1\linewidth]{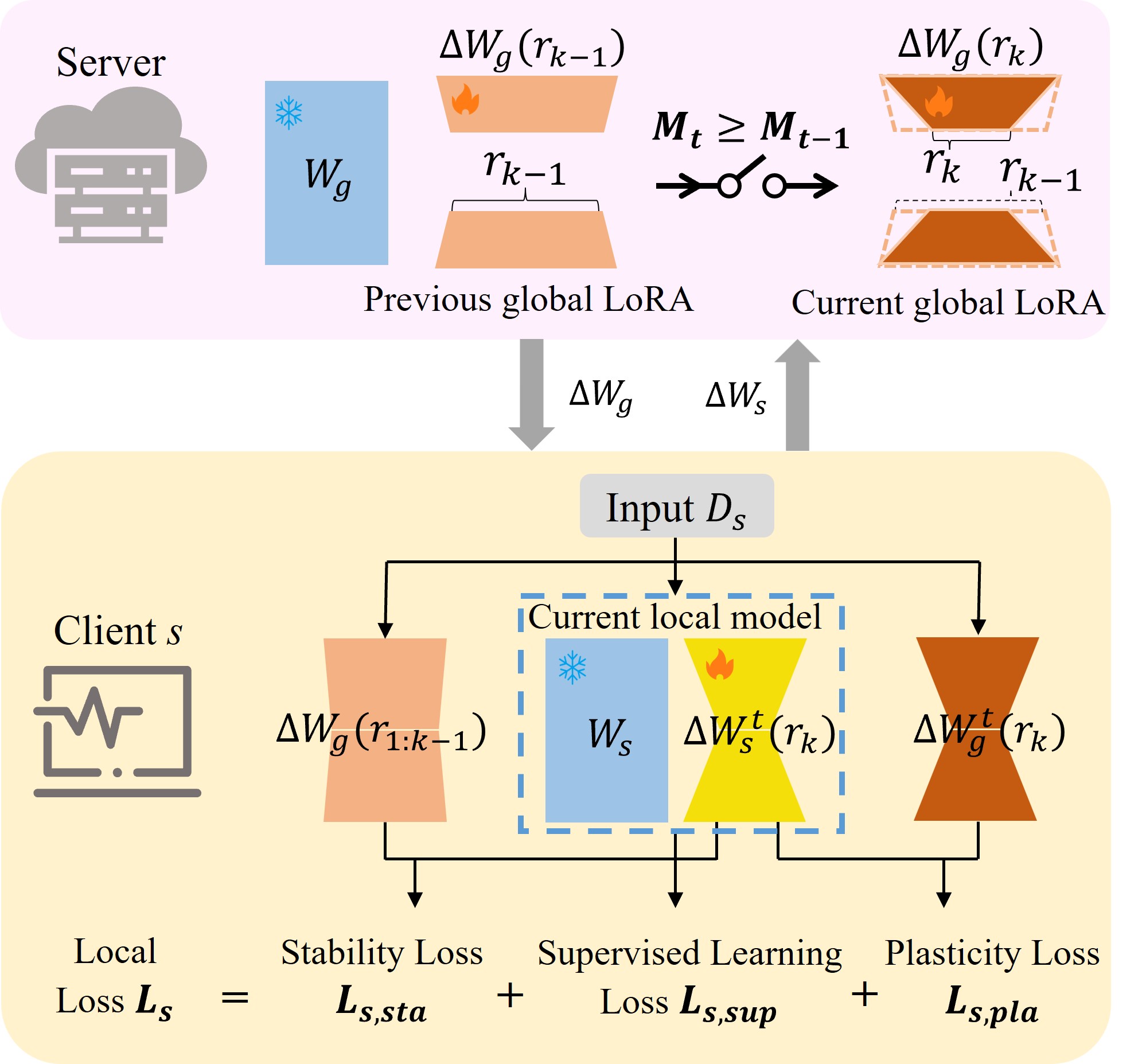}
    \caption{An overview of \methodname{}.}
    \label{fig:overview}
\end{figure}

As shown in Figure \ref{fig:overview}, our paper proposes the \methodname{} framework, which gradually reduces the rank of LoRA modules to find the most appropriate parameter dropout rate for a specific level of model performance. There are $\mathcal{S}=\{1,2,\dots, S\}$ clients and a server (denoted as $g$). All clients have the same fixed pre-trained weight matrix $W_s$ and three low-rank decomposition matrices (one local LoRA module and two global LoRA modules with different dropout rates). The server aggregates the collected local LoRA modules in the global aggregation phase. \textit{i)} On the server, we design a gradient-based stepwise parameter dropout mechanism to reduce the communication cost, which gradually drops the LoRA parameters. We use SGC value ($M^t$) to reflect the convergence stability of the global model. Once SGC stops decreasing (i.e., $M^t \geq M^{t-1}$), we reduce the LoRA rank by one until reaching the minimum rank value ($r^k < r^{k-1}$). \textit{ii)} On the local clients, we introduce CL to tackle the performance degradation resulting from the inconsistent optimal objectives of varying parameter dropout rates. The local loss consists of three parts: the supervised learning loss $L_{s,sup}$, the stability loss $L_{s,sta}$ and the plasticity loss $L_{s,pla}$. The stability and plasticity losses are derived by alignment of the current local LoRA module $\Delta W_s^t(r_k)$ with the accumulated previous global LoRA module $\Delta W_g(r_{1:k-1})$ and the current global LoRA module $\Delta W_g^t(r_k)$, respectively. 

To illustrate the workflow of \methodname{}, we first introduce the relevant preliminary, then design the parameter dropout mechanism and local training algorithm respectively, and finally give a detailed complexity analysis.

\subsection{FL with LoRA Tuning}

LoRA is a typical reparameterization-based method of PEFT with no inference latency. As shown in Figure \ref{fig:LoRA}, the low-rank decomposition matrix (LoRA module) $\Delta W=BA$ is injected alongside the pre-trained weight matrix $W \in \mathbb{R}^{h_1 \times h_2}$, where $B \in \mathbb{R}^{h_1 \times r}$ is initialized with zero and $A \in \mathbb{R}^{r \times h_2}$ with a random independent Gaussian coefficients (rank $r \ll h_1,h_2$). Note that the trainable parameter dropout rate is controlled by the rank $r$ and the original dimension $h_1,h_2$. Thus, the forward pass yields:
\begin{equation}
    \hat{y}=(W+\Delta W)x = Wx+BAx,
\end{equation}
where the output vectors of $Wx$ and $BAx$ are summed coordinate-wise.

We denote the local dataset at client $s$ as $\mathcal{D}_s=\{(x,y) \vert x\in\mathcal{X},y \in \mathcal{Y}\}$, where $\mathcal{X}$ and $\mathcal{Y}$ are the feature space and the label space, respectively. There are $\mathcal{E}=\{1,2,...,E\}$ local training epochs in each communication round and $\mathcal{T}=\{1,2,...,T\}$ communication rounds. 

For the local training phase at client $s$, the local objective function $L_s^{t,e}$ in epoch $e$ of round $t$ is the supervised learning loss $L_{s,sup}^{t,e}$ under the local model weights $W_s^{t,e}+\Delta W _s^{t,e}$:

\begin{equation}
\label{equ:localloraloss}
\begin{aligned}
    L_s^{t,e}=L_{s,sup}^{t,e}=\mathbb{E}_{(x,y) \sim \mathcal{D}_s}[\mathcal{L}_s(x,y;W_s^{t,e}+\Delta W _s^{t,e})],\\ \forall s\in\mathcal{S}, e\in\mathcal{E}, t\in\mathcal{T},
\end{aligned}
\end{equation}
where the cross-entropy loss is typically used for $\mathcal{L}_s(\cdot)$.

With the frozen pre-trained model and trainable LoRA modules, client $s$ computes local loss gradients and updates the LoRA modules via backpropagation with learning rate $\eta$:
\begin{equation} \label{equ:update}
    \Delta W_s^{t,e} \leftarrow \Delta W_s^{t,e-1} - \eta \nabla_{\Delta W_s} L_s^{t,e-1}, \forall s\in\mathcal{S}, e\in\mathcal{E}, t\in\mathcal{T}.
\end{equation}
Particularly, the weight $\Delta W_s^{t,E}$ in the last local epoch is abbreviated as $\Delta W_s^t$.

For the global aggregation phase, the server aggregates all local LoRA modules across $S$ clients by weighted averaging to generate a new global LoRA module:
\begin{equation} \label{equ:global_lora}
   \Delta W_g^t = \sum_{s\in\mathcal{S}} \frac{|\mathcal{D}_s|}{|\mathcal{D}|} \Delta W_s^t, \forall t\in\mathcal{T},
\end{equation}
where $\mathcal{D} \triangleq \bigcup_{s\in\mathcal{S}}\mathcal{D}_s$ represents the union of local datasets. The global LoRA modules will be distributed to all local clients to initialize the local LoRA modules for the next round.

The global loss $L_g$ is the weighted average of local losses $L_s$. The final global optimization objective is to find a group of global LoRA parameters that minimize the global loss:
\begin{equation} \label{equ:global_loss}
    \mathop{\arg\min} \limits_{\Delta W_g} L_g=\sum_{s\in\mathcal{S}} \frac{|\mathcal{D}_s|}{|\mathcal{D}|} L_s.
\end{equation}

\subsection{Server Side Design}

In this subsection, we propose a gradient-based stepwise parameter dropout mechanism that gradually increases the dropout rate to reduce communication costs on the server side. This mechanism uses a novel metric, SGC, to measure the training status.
Previous research \cite{chen2023gift} has proved that the gradient consistency is able to reflect the training status of FL. At the beginning of FL, local models across clients uniformly update towards the global optimal objective. As the objective function approaches the global optimum, they gradually exhibit a bifurcating trend. Therefore, the gradient consistency continuously decreases as the federated training iterates until the training stabilizes. In our method, the parameter dropout rate will rise to find a new optimization space when the training stabilizes and stagnates. 
The detailed process of stepwise parameter dropout on the server side is presented in Algorithm \ref{alg:server}. 

\begin{algorithm} [t]
    \caption{Parameter Dropout on the Server.} 
    \label{alg:server}
    \KwIn{Global LoRA parameter $\Delta W_g^t$; local LoRA parameter $\Delta W_s^t$; decay factor $\theta$ and $\lambda$; subtractor $\delta$.} 
    \KwOut{The accumulated global LoRA module $\Delta W_g(r_{1:k})$.} 
        Get sensitivity $U_s^t$ from the accumulated gradient $grad(\Delta W_s^t)$ based on equation (\ref{equ:sen});  \\
        Compute the normalization of sensitivity $\alpha_s^t$ from $U_s^t$ by equation (\ref{equ:sen_w});  \\
        Calculate EMA $\Bar{P}^t$ and $\Bar{N}^t$ from $grad(\Delta W_s^t)$ and $\alpha_s^t$ based on equations (\ref{equ:pg})-(\ref{equ:eng});  \\
        Get \textit{SGC} $M^t$ by equation (\ref{equ:sgc});  \\
        \If{$M^t \geq M^{t-1}$} 
            {Execute \textit{Parameter Dropout}: $r_k \leftarrow r_{k-1} - \delta$; \\
            Accumulate global LoRA $\Delta W_g(r_{1:k})$ based on equation (\ref{equ:accLoRA}).}
\end{algorithm}

\textbf{Sensitivity:} Firstly, we adopt the real-time sensitivity to reveal the influence of client parameters \cite{zhang2022platon,zhang2023adalora,zhang2023increlora}. Let $grad(\Delta W_s^t) = \Delta W_s^t - \Delta W_g^t $ be the accumulated gradient of client $s$ in $t$-th round. The sensitivity of LoRA is defined as the magnitude of the gradient-parameter product:
\begin{equation} \label{equ:sen}
    U_s^t = |\Delta W_s^t \cdot grad(\Delta W_s^t)|, \forall s\in\mathcal{S}, t\in\mathcal{T},
\end{equation}
where a high sensitivity value means a significant impact on loss when the parameters are removed.

We further normalize the sensitivity as the weight of client $s$:
\begin{equation} \label{equ:sen_w}
    \alpha_s^t = \frac{U_s^t}{\sum_{s\in\mathcal{S}} U_s^t}, \forall t\in\mathcal{T},
\end{equation}
where $\alpha_s^t$ meets the constraint $\sum_{s=1}^S \alpha_s^t =1$.

\textbf{SGC:} To obtain real-time gradient statistics stably, we propose sensitivity-based bilateral gradient pooling, which collects both positive and negative components from all clients. The divergence of local gradients across clients also leads to the separation of positive and negative components.

We apply the $\bm{Relu}$ function to $grad(\Delta W_s^t)$, a classical sign filtering tool. The sensitivity-based positive gradient component is defined as:
\begin{equation} \label{equ:pg}
    P^t = \sum_{s\in\mathcal{S}} \alpha_s^t \cdot \bm{Relu} \left(grad(\Delta W_s^t)\right), \forall t\in\mathcal{T}.
\end{equation}

Similarly, the sensitivity-based negative gradient component is:
\begin{equation} \label{equ:ng}
    N^t = \sum_{s\in\mathcal{S}} -\alpha_s^t \cdot \bm{Relu} \left(-grad(\Delta W_s^t)\right), \forall t\in\mathcal{T}.
\end{equation}

To further improve memory efficiency, we use the gradients’ exponential moving average (EMA) to smooth the positive gradient components with historical records. Thus, EMA with the decay factor $\theta$ of the positive gradient component is derived as:
\begin{equation} \label{equ:epg}
    \Bar{P}^t = \langle P^t \rangle _\theta = \theta \times \Bar{P}^{t-1} + (1-\theta) \times P^t, \forall t\in\mathcal{T}.
\end{equation}

EMA of the negative gradient component is:
\begin{equation} \label{equ:eng}
    \Bar{N}^t = \langle N^t \rangle _\theta = \theta \times \Bar{N}^{t-1} + (1-\theta) \times N^t, \forall t\in\mathcal{T}.
\end{equation}

Finally, SGC is formulated as
\begin{equation} \label{equ:sgc}
    M^t = \frac{\|\Bar{P}^t+\Bar{N}^t\|}{\|\Bar{P}^t\| + \|\Bar{N}^t\|}, \forall t\in\mathcal{T},
\end{equation}
where $M^t$ is 1 when gradients of all clients have a strong consistency at FL commencement, and decreases to 0 as FL iterates until they counteract with each other. However, this is extremely ideal which is not realistic in FL settings due to the influence of random mini-batches. Therefore, we view the comparison of SGC in the adjacent two rounds as the feedback signal instead of the absolute value of SGC.

When FL training stabilizes and stagnates, i.e., $M^t \geq M^{t-1}$, \methodname{} scales up the parameter dropout rate by a fixed subtractor $\delta$ of LoRA rank. 
\begin{equation}
    r_k \leftarrow r_{k-1} - \delta, \forall k\in\mathcal{K}=\{1,2,...,K\}.
\end{equation}

We abbreviate LoRA $\Delta W$ of the final round under the rank $r_k$ as $\Delta W(r_k)$. The accumulated LoRA with the previous rank under the decay factor $\lambda$ is denoted as:
\begin{equation} \label{equ:accLoRA}
    \Delta W_g(r_{1:k}) = \lambda \times \Delta W_g(r_{1:k-1}) + (1-\lambda) \times \Delta W_g(r_{k}), \forall k\in\mathcal{K}.
\end{equation}

\subsection{Client Side Design}

In this subsection, we introduce CL to FL to trade off the stability and plasticity, where the performance degradation results from the varying parameter dropout rate. The proposed continual federated learning on the client side is shown in Algorithm \ref{alg:client}.

Due to the inconsistent optimal objectives under different parameter dropout rates, the model inevitably suffers from catastrophic forgetting, with performance degradation occurring after optimization with fewer trainable parameters. CL helps maintain previous knowledge to guide the learning of current data. Therefore, to mitigate the negative impact of varying parameter dropout rates in non-IID FL, \methodname{} initializes the global model using the accumulated previous parameters $\Delta W_g(r_{1:k-1})$ and optimizes the current local model $\Delta W_s^{t,e}(r_k)$ by continually learning from the previous parameters, rectifying the update directions of local models towards the global one with the regularization strength $\mu_1$. Thus, the regularization term of stability is derived as:
\begin{equation} \label{equ:loss_sta}
\begin{aligned}
    L_{s,sta}^{t,e} = \Theta\left(\Delta W_s^{t,e}(r_k);\Delta W_g(r_{1:k-1}),\mu_1\right), \\
    \forall s\in\mathcal{S}, e\in\mathcal{E}, t\in\mathcal{T}, k\in\mathcal{K},
\end{aligned}
\end{equation}
where $\Theta(\cdot)$ is the function of regularization.

For example, Elastic Weight Consolidation (EWC) \cite{kirkpatrick2017overcoming} approximates the function using the Fisher Information Matrix (FIM): $\Theta(W;W^{\ast},\mu) = \frac{\mu}{2} \sum_{j\in\mathcal{J}} F_j(W_j-W_j^{\ast})^2$, where $\mathcal{J}=\{1,2,\dots,J\}$ denotes the network layers. Memory Aware Synapses (MAS) \cite{aljundi2018memory} accumulates parameter changes based on the magnitude of updates. Learning without Forgetting (LwF) \cite{li2017learning} learns the soft target and minimizes the gap between the logits of the previous and current models. For the sake of simplicity, these three regularization-based CL methods are summarized in Table \ref{tab:reg_method}.

\begin{algorithm}[t]
    \caption{Local Training with CL on the Clients.} 
    \label{alg:client}
    \KwIn{Datasets from $S$ clients: $\mathcal{D}_1$, $\mathcal{D}_2$, \ldots, $\mathcal{D}_S$; local epochs $E$; learning rate $\eta$; regularization strength $\mu_1$, $\mu_2$.} 
    \KwOut{The local LoRA module $\Delta W_s^t(r_k)$.} 
        \For{$e=1, 2, \ldots, E$}
        {
            Get the supervised learning Loss $L_{s,sup}^{t,e}$ by equation (\ref{equ:localloraloss}); \\
            Compute the \textit{Stability} loss $L_{s,sta}^{t,e}$ based on equation (\ref{equ:loss_sta}); \\
            Calculate the \textit{Plasticity} loss $L_{s,pla}^{t,e}$ by equation (\ref{equ:loss_pla}); \\
            Get the local loss $L_s^{t,e}$ by adding $L_{s,sup}^{t,e}$, $L_{s,sta}^{t,e}$ and $L_{s,pla}^{t,e}$ based on equation (\ref{equ:local_loss}); \\
            Update the local LoRA module $W_s^{t,e}(r_k)$ by equation (\ref{equ:update});
        }
        $\Delta W_s^t(r_k) \leftarrow \Delta W_s^{t,E}(r_k)$.
\end{algorithm}

\begin{table}[t]
\scriptsize
    \caption{The regularization function $\Theta(\cdot)$ of continual learning.}
    \label{tab:reg_method}
    \centering
    \begin{tabular}{c|c}
    \toprule
    CL Method   &  $\Theta(W;W^{\ast},\mu)$  \\ 
    \hline
    EWC      &  $\frac{\mu}{2} \sum_j F_j(W_j-W_j^{\ast})^2$   \\ 
    MAS      &  $\frac{\mu}{2} \sum_j M_j(W_j-W_j^{\ast})^2$    \\
    LwF      &  $\mu \sum_{c=1}^{C_{1:t}} -y^c(x;W^{\ast})log(y^c(x;W))$    \\
    \bottomrule
    \end{tabular}
\end{table}

Although the old parameters are beneficial for model stability, they limit the model’s ability to learn new knowledge. When the optimal direction of new parameters diverges from the previous one, it disrupts the balance between stability and plasticity. Similarly, \methodname{} utilizes the current global model with a new parameter dropout rate $\Delta W_g^t(r_k)$ to constantly correct the local training, which aims to maximize the consistency between the current local and global models and avoid local overfitting. Like equation (\ref{equ:loss_sta}), we define the regularization term of plasticity with the regularization strength $\mu_2$ as:
\begin{equation} \label{equ:loss_pla}
\begin{aligned}
    L_{s,pla}^{t,e} = \Theta\left(\Delta W_s^{t,e}(r_k);\Delta W_g^t(r_k),\mu_2\right), \\
    \forall s\in\mathcal{S}, e\in\mathcal{E}, t\in\mathcal{T}, k\in\mathcal{K}.
\end{aligned}
\end{equation}

The regularization term of stability binds the dynamic of the local parameters to the old accumulated global parameters, while the regularization term of plasticity boosts the local model to learn new global optima. Thus, the local loss of equation (\ref{equ:localloraloss}) can be deduced by:
\begin{equation} \label{equ:local_loss}
    L_s^{t,e} = L_{s,sup}^{t,e} + L_{s,sta}^{t,e} + L_{s,pla}^{t,e}, \forall s\in\mathcal{S}, e\in\mathcal{E}, t\in\mathcal{T}.
\end{equation}

Through appropriate adjustment of regularization strengths $\mu_1$ and $\mu_2$. \methodname{} realizes a better trade-off of stability and plasticity, mitigating the performance degradation caused by the parameter dropout. The detailed procedure of \methodname{} is outlined in Algorithm \ref{alg:all} (consisting of Algorithm \ref{alg:server} and \ref{alg:client}).

\subsection{Computational Complexity}

This subsection discusses the computational requirements of three algorithms, which is helpful for understanding scalability and efficiency. The number of multiplication operations in the LoRA module is represented as $\psi$. Additionally, $\gamma$ represents the mini-batch size. Thus, the model has the $\psi\gamma$ trainable parameters. Regarding the complexity of Algorithm \ref{alg:server}, since the deduction of SGC conducts the same operations for each gradient coordinate, the computational complexity can be expressed as $O(\psi\gamma)$. Regarding Algorithm \ref{alg:client}, the computational complexity of backpropagation for each local training iteration is straightforwardly $O(\psi\gamma)$. Therefore, the computational complexity of Algorithm \ref{alg:client} with $E$ epoches is $O(E\psi\gamma)$. To determine the computational complexity of Algorithm \ref{alg:all}, we utilize the analysis of Algorithm \ref{alg:server} and \ref{alg:client}, deriving that the computational complexity of \methodname{} is $O(TSE\psi\gamma)$.

\section{Experimental Evaluation}

\subsection{Experiment Setup}

\subsubsection{Datasets and Models}

To validate the efficiency of \methodname{}, we conduct experiments on the public CIFAR-10 dataset and a private healthcare dataset of Face. The CIFAR-10 dataset is evaluated on the Swin-T model from the Swin Transformer family \cite{liu2021swin}. Following the data partition of the previous study \cite{qu2022rethinking}, we adopt the original testing set with 10,000 images as the global test dataset, randomly select 5,000 images from the original training dataset as the global validation dataset, and manually partition the remaining 45,000 images as the training dataset into 5 clients. The CIFAR-10 dataset uses the mean Kolmogorov Smirnov (KS $\in[0,1]$) statistic between two clients to control the label skew. An increase in the KS value from 0 to 1 indicates that the skew degree varies from the independent identically distributed (IID) to an extremely non-IID. `Split\_1' is IID with KS=0. `Split\_2' is moderate non-IID with KS=0.65. Specifically, each client samples images from four classes, two of which overlap with other clients. `Split\_3' is the extreme non-IID with KS=1 where each client receives two disjoint classes.

To further validate our FL method in more challenging settings, the medical Face dataset is trained on the Query2Label model with multi-label disease tasks, including 7 labels of diseases (AF, Angina, Gout, Hyperthyroidism, Hypothyroidism, Malignancy Colorectum, Malignancy Ovary). Each disease label is binarized into positive and negative categories. We collect 1,000 samples as the global test set, and each of the four clients, based on the hospital locations, retains 5,000 images with a training and validation ratio of 3:1. For a fair comparison, the models used in this paper are pre-trained on ImageNet \cite{deng2009imagenet} according to algorithm requirements. 

\subsubsection{Evaluation Metrics}

Experiments adopt the 5-fold cross-validation method to assess and test model performances. We evaluated the models’ performance by test accuracy (ACC) of the CIFAR-10 dataset and Area-Under-Curve (AUC) of the Face dataset. Since TP, TN, FP and FN act as the number of true positive, false negative, false positive and false negative, respectively, ACC is the true prediction ratio:
\begin{equation}
    ACC=\frac{TP+TN}{TP+TN+FP+FN}.
\end{equation}
AUC is the area under the ROC curve, whose X-axis is FPR = FP/(FP+TN) and Y-axis is TPR = TP/(TP+FN).

\begin{algorithm} [t]
    \caption{\methodname{}.} 
    \label{alg:all}
    \KwIn{Datasets from $S$ clients: $D_1$, $D_2$, \ldots, $D_S$; communication rounds $T$; learning rate $\eta$; rank $r_K$; decay factor $\theta$ and $\lambda$; subtractor $\delta$; regularization strength $\mu_1$, $\mu_2$.} 
    \KwOut{The final global LoRA parameter $\Delta W_g$.} 
        Initialization: $W_g$, $\Delta W_g^1(r_1)$\; 
        \For{$t=1, 2, \ldots, T$} 
        { 
            Decide whether to drop parameters by Algorithm \ref{alg:server};  \\
            \For{$s=1, 2, \ldots, S$} 
            {
                Distribute global LoRA module $\Delta W_g^t(r_k)$ to client $s$; \\
                Compute the updated local LoRA module $\Delta W_s^t(r_k)$ by Algorithm \ref{alg:client}; \\
                Upload the local LoRA module $\Delta W_s^t(r_k)$ to the server.\\
            }
             Get and save the global LoRA module $\Delta W_g^{t}$ based on equation (\ref{equ:global_lora}); \\
             Calculate the global loss $L_g^t$ by equation (\ref{equ:global_loss}).
        } 
\end{algorithm}

When the global model achieves the highest ACC or AUC, the LoRA rank is regarded as the optimal $r_k$. We compute the transmitted parameters until this time as the communication cost:
\begin{equation}
    Communication\ Cost = 2 \times S \times \sum_{t\in\mathcal{T}} Param^t,
\end{equation}
where $Param^t$ is the trainable parameters transmitted between the server and each client in $t$-th round.

To investigate the effects of \methodname{} on the stability-plasticity dilemma, we further employ two metrics: Weight Distance (WD) and Centered Kernel Alignment (CKA) \cite{kim2023achieving}. Based on Taylor expansion, WD is determined by the norm of the difference between the weights of the local LoRA module $\Delta W_s^t(r_k)$ with the accumulated previous global LoRA module $\Delta W_g(r_{1:k-1})$ and the current global LoRA module $\Delta W_g^t(r_k)$ respectively:
\begin{equation}
    WD_{s,sta}^t = {\Vert \Delta W_s^t(r_k) - \Delta W_g(r_{1:k-1}) \Vert}_2,
\end{equation}
\begin{equation}
\begin{aligned}
    WD_{s,pla}^t = {\Vert \Delta W_s^t(r_k) - \Delta W_g^t(r_k) \Vert}_2, \\
    \forall s\in\mathcal{S}, t\in\mathcal{T}, k\in\mathcal{K}.
\end{aligned}
\end{equation}
CKA is used to measure the similarity of two representations $Z$, which is implemented by the linear Hilbert-Schmidt Independence Criterion (HSIC): 
\begin{equation}
    CKA(Z_1, Z_2) = \frac{HSIC(Z_1, Z_2)}{\sqrt{HSIC(Z_1, Z_1)} \times \sqrt{HSIC(Z_2, Z_2)}}.
\end{equation}
In our paper, we calculate an average CKAs over the total $J$ layers of the LoRA module. CKAs of stability and plasticity are defined as follows:
\begin{equation}
    CKA_{s,sta}^t = \frac{1}{J} \sum_{j\in\mathcal{J}}CKA(Z_{s,j}^t(r_k), Z_{g,j}(r_{1:k-1})),
\end{equation}
\begin{equation}
\begin{aligned}
    CKA_{s,pla}^t = \frac{1}{J} \sum_{j\in\mathcal{J}}CKA(Z_{s,j}^t(r_k), Z_{g,j}^t(r_k)), \\
    \forall s\in\mathcal{S}, t\in\mathcal{T}, k\in\mathcal{K},
\end{aligned}
\end{equation}
where $Z_{s,j}^t(r_k)$, $Z_{g,j}(r_{1:k-1})$ and $Z_{g,j}^t(r_k)$ are the representations learned from the $j$-th layer of $\Delta W_s^t(r_k)$, $\Delta W_g(r_{1:k-1})$ and $\Delta W_g^t(r_k)$, respectively.

\begin{table*}[t]
\renewcommand{\arraystretch}{1}
\caption{Accuracy and communication efficiency of different algorithms on CIFAR-10 dataset.}
\label{tab:CIFAR-10}
\centering
    \begin{tabular}{|l|cr|cr|cr|}
    \hline
    \multirow{2}{*}{Methods} & \multicolumn{2}{c}{Split\_1} & \multicolumn{2}{|c|}{Split\_2} & \multicolumn{2}{c|}{Split\_3}  \\ 
    \cline{2-7} 
      & ACC$\uparrow$  & \makecell{Communication \\ Cost (MB)}$\downarrow$  & ACC$\uparrow$  & \makecell{Communication \\ Cost (MB)}$\downarrow$  & ACC$\uparrow$  & \makecell{Communication \\ Cost (MB)}$\downarrow$     \\ 
    \hline
    FedAvg   & \textbf{0.9505}  & 171161.9626  & 0.9270  & 153310.7149  & 0.7235  & 187963.1368 \\
    FedGLF   & 0.9420  & 160288.7962  & 0.9070  & 164485.5066  & 0.7625  & 165028.1456 \\
    FedBug   & 0.7840  &  72287.8253  & 0.7015  &  73337.8987  & 0.4575  &  74387.9721 \\
    FedPEFT-Bias  & 0.8265  &  \textbf{438.0990}  & 0.7540  &  \textbf{389.1010}  & 0.4145  &  \textbf{377.5721} \\
    FedPEFT-LoRA  & 0.9500  &    857.1102  & 0.9325  &    792.4226  & 0.8850  &    948.7509 \\
    FedPEFT-LoRA-FA  & 0.8910  &    498.6328  & 0.7740  &    522.8901   & 0.4520   &    525.5855  \\
    SLoRA   & 0.9440   &    716.4038  & 0.9255  &    729.8804  & 0.8515  &    781.0913 \\
    \hline
    \textbf{\methodname{}} (EWC)  & \textbf{0.9505} &    787.0314  & \textbf{0.9350}  &    708.8673  & \textbf{0.8880}  &    711.5625 \\
    \textbf{\methodname{}} (MAS)  & 0.9490  &  774.2288  & 0.9340  &    741.8848  & 0.8860 	 &    747.2755 \\
    \textbf{\methodname{}} (LwF)  & \textbf{0.9505} &    727.0607  & 0.9325  &    696.0646  & 0.8875  &    718.3008 \\
    \textbf{\methodname{}} (Average)  & 0.9500 &    762.7736  & 0.9338  &    715.6056  & 0.8872  &    725.7129 \\
    \hline
    \end{tabular}
\end{table*}

\subsubsection{Comparison Baselines}

We compare \methodname{} with six state-of-the-art baseline methods. The first three methods are communication-efficient FL without PEFT, while the remaining three incorporate PEFT into FL:
\begin{enumerate}
    \item \textbf{FedAvg} \cite{mcmahan2017communication}: The entire local model parameters are updated and shared;
    \item \textbf{FedGLF} \cite{malan2022communication}: The local models gradually freeze from the shallower to deeper layers and share the unfreezing layers; 
    \item \textbf{FedBug} \cite{kao2023fedbug}: On the contrary to FedGLF, FedBug gradually unfreezes from the bottom to up layers and transmits the unfreezing layers;
    \item \textbf{FedPEFT-Bias}: Clients train and update local pre-trained models with the BitFit method \cite{zaken2021bitfit}, and only share the bias vectors while freezing everything else;
    \item \textbf{FedPEFT-LoRA}: Clients train and update the additional LoRA module \cite{hulora} with fixed rank of local pre-trained models, and only share the LoRA vectors while freezing everything else.
    \item \textbf{FedPEFT-LoRA-FA} \cite{zhang2023lora}: Clients freeze the projection-down weight $A$ while updating and sharing the projection-up weight $B$ of LoRA.
    \item \textbf{SLoRA} \cite{babakniya2023slora}: Stage 1 adopts sparse fine-tuning by generating a random data-independent binary mask with uniform density. Then stage 2 follows the standard FedPEFT-LoRA with fixed rank.
\end{enumerate}

\subsubsection{Implementation Details}

Experiments were conducted using Pytorch 2.1.0 with Python 3.11 on NVIDIA A40 GPUs. We tuned the following hyper-parameters and report the best results. For the CIFAR-10 dataset, we chose an SGD optimizer with an initial learning rate of 0.003 and a cosine learning rate scheduler for the FedAvg, FedGLF and FedBug, while the initial learning rate of other algorithms was set to 0.02. The maximal communication round of $T$ was set to 200 and the local epoch $E$ was set to 1, with all local clients participating in FL training in each round. The initial rank $r_1$, the final rank $r_K$ and subtractor $\delta$ are set to 16, 8 and 2, respectively. For the Face dataset, we used an Adam optimizer with an initial learning rate of 0.0001 and a cosine learning rate scheduler for all methods. The number of local training epochs $E$ on each client was set to 2 and the total number of communication rounds $T$ to 50. 

\begin{table*}[ht]
\renewcommand{\arraystretch}{1}
\caption{Accuracy and communication efficiency of different algorithms on the Face dataset.}
\label{tab:Face}
\centering
\resizebox*{1\linewidth}{!}{
    \begin{tabular}{|l|cccccccc|r|}
    \hline
    \multirow{2}{*}{Methods} & \multicolumn{8}{c|}{AUC$\uparrow$} & \multirow{2}{*}{\makecell{Communication \\ Cost (MB)}$\downarrow$} \\ 
    \cline{2-9}
      &  AF  &  Angina  &  Gout  &  Hyperthyroidism  &  Hypothyroidism  &  \makecell[c]{Malignancy \\ Colorectum}  &  \makecell[c]{Malignancy \\ Ovary}  & Average  &                   \\ 
    \hline
    FedAvg   &  0.6700  &  \textbf{0.6500}  &  0.6600  &  0.5400  &  \textbf{0.8600}  &  0.2400  &  0.7000  &  0.6171  &  37228.3264  \\
    FedGLF   &  0.6432  &  0.6340  &  0.6824  &  0.3030  &  0.8340  &  0.5616  &  0.8007  &  0.6370  &  14260.0330  \\
    FedBug   &  0.5084  &  0.6108  &  0.5810  &  0.5000  &  0.4600  &  0.3082  &  0.4825  &  0.4930  &  28420.4745  \\
    FedPEFT-Bias  &  0.4497  &  0.3667  &  0.6916  &  \textbf{0.8496}  &  0.2780  &  0.0890  &  0.5385  &  0.4662  &  75.7614     \\
    FedPEFT-LoRA  &  0.6718  &  0.6192  &  0.7942  &  0.6801  &  0.7376  &  0.7603  &  \textbf{0.8217}  &  0.7264  &  5.7500   \\
    FedPEFT-LoRA-FA  &  0.5481  &  0.3196  &  0.6420  &  0.7055  &  0.0511  &  0.2534  &  0.3916 &  0.4159  &  \textbf{2.2500}   \\
    SLoRA    &  \textbf{0.7008}  &  0.6481  &  0.7221  &  0.4237  &  0.8355  &  0.7260  &  0.7465  &  0.6861  &  1179.3258    \\ 
    \hline
    \textbf{\methodname{}} (EWC) &  0.6687  &  0.6256  &  \textbf{0.7962}  &  0.6949  &  0.7645  &  \textbf{0.8288}  &  0.8129  &  0.7417  &  4.0625    \\
    \textbf{\methodname{}} (MAS) &  0.6683  &  0.6252  &  \textbf{0.7962}  &  0.6949  &  0.7674  &  \textbf{0.8288}  &  0.8147  &  \textbf{0.7422}  &  3.9063    \\
    \textbf{\methodname{}} (LwF) &  0.6684  &  0.6245  &  \textbf{0.7962}  &  0.6907  &  0.7603  &  \textbf{0.8288}  &  0.8129  &  0.7403  &  4.1875    \\ 
    \textbf{\methodname{}} (Average) &  0.6685  &  0.6251  &  \textbf{0.7962}  &  0.6935  &  0.7641  &  \textbf{0.8288}  &  0.8135  &  0.7414  &  4.0521    \\ 
    \hline
    \end{tabular}
}
\end{table*}

\begin{table*}[ht]
\renewcommand{\arraystretch}{1}
\caption{Effects of different CL methods.}
\label{tab:abl_CL}
\centering
\resizebox*{1\linewidth}{!}{
    \begin{tabular}{|c|cc|cr|cr|cr|}
    \hline
    \multirow{2}{*}{CL Methods} & \multirow{2}{*}{Stability} & \multirow{2}{*}{Plasticity} & \multicolumn{2}{c|}{Split\_1} & \multicolumn{2}{c|}{Split\_2} & \multicolumn{2}{c|}{Split\_3} \\ 
    \cline{4-9} 
     & & & ACC$\uparrow$  & \makecell{Communication \\ Cost (MB)}$\downarrow$  & ACC$\uparrow$ & \makecell{Communication \\ Cost (MB)}$\downarrow$  & ACC$\uparrow$  & \makecell{Communication \\ Cost (MB)}$\downarrow$ \\ 
    \hline
    \textbf{\methodname{}} w/o CL  &  -  &  -  &  0.9475  &  776.9239  &  0.9320  &  692.0216  &  0.8840   &  669.1114  \\ 
    \hline
    \multirow{3}{*}{\textbf{\methodname{}} (EWC)} &  $\surd$  &  -  &  0.9475  &  751.9924  &  0.9330  &  758.7306  &  \textbf{0.8880}  &  \textbf{646.8750}  \\
      &  -  &  $\surd$  &  0.9435  &  \textbf{552.5394}  &  0.9340  &  747.9494  &  0.8830  &  711.5626  \\
      &  $\surd$  &  $\surd$  &  \textbf{0.9505}  &  787.0314  &  \textbf{0.9350}  &  \textbf{708.8673}  &  \textbf{0.8880}   &  711.5625  \\ 
    \hline
    \multirow{3}{*}{\textbf{\methodname{}} (MAS)} &  $\surd$  &  -  &  0.9460  &  731.1037  &  0.9330  &  754.6876  &  0.8860  &  737.8419  \\
      &  -  &  $\surd$  &  0.9430  &  \textbf{586.9047}  &  \textbf{0.9340}  &  \textbf{723.0176}  &  0.8760  &  \textbf{705.4983}  \\
       &  $\surd$  &  $\surd$  &  \textbf{0.9490}  &  774.2288  &   \textbf{0.9340}  &  741.8848  &  \textbf{0.8900}  &  747.2755  \\ 
    \hline
    \multirow{3}{*}{\textbf{\methodname{}} (LwF)} &  $\surd$  &  -  &  0.9470  &  783.6623  &  0.9320  &  714.9318  &  0.8845  &  731.7773  \\
      &  -  &  $\surd$  &  \textbf{0.9510}    &  735.8205  &  \textbf{0.9350}  &  744.5802  &  0.8830  &  \textbf{691.3477}  \\
       &  $\surd$  &  $\surd$  &  0.9505  &  \textbf{727.0607}  &  0.9325  &  \textbf{696.0646}  &  \textbf{0.8875}  &  718.3008  \\ 
    \hline
    \end{tabular}
}
\end{table*}

\subsection{Results and Discussion}

\subsubsection{Performances on the CIFAR-10 Dataset}

Our study conducts experiments on the natural CIFAR-10 dataset with three non-IID settings to verify the efficiency and effectiveness of \methodname{}, which operates with three different CL types. As the main results in Table \ref{tab:CIFAR-10} show, FedAvg transmits 100\% of the model parameters to reach convergence, leading to significantly overloaded communication. Communication-efficient FL methods, such as FedGLF and FedBug, reduce a tiny communication burden due to their maintenance of pre-trained models. FedBug performs particularly poorly because it wastes too many resources on the shallow-level features at the commencement, which does not facilitate model convergence. In contrast, PEFT methods perform better in terms of communication efficiency. Although FedPEFT-Bias tunes the bias terms with different levels and consumes minimal communication costs, its accuracy is still inferior to that of FedPEFT with LoRA, such as SLoRA, FedPEFT-LoRA-FA and the best-performing FedPEFT-LoRA.

Our paper aims at improving the communication efficiency guided by a satisfactory model performance, so result analyses focus on the comparison of SPD-CFL and the baseline with the best test ACC. Compared to the above approaches, \methodname{} achieves comparable test accuracy with significantly lower parameter transmission, benefiting from its runtime adaptability and consistency. While most algorithms exhibit negligible performance differences in the IID (Split\_1) setting, only \methodname{} (EWC) and \methodname{} (LwF) achieve accuracy comparable to the classical FedAvg while reducing the average communication cost by 99.55\%. By aligning the local LoRA module with the global LoRA module under varying parameter dropout rates, SPD-CFL corrects the local optimization direction towards the global optima, thereby mitigating the FL non-IID challenge and improving global generalization. Even in the extremely non-IID (Split\_3) setting, \methodname{} stands out with the highest test accuracy among all efficient FL methods and reduces communication burden by 23.51\% compared to the best-performing existing work FedPEFT-LoRA. In addition, the non-IID data seriously degrades model performance due to the highly inconsistent local and global optimization objectives. However, while the average accuracy of \methodname{} decreases by only 6.61\% from Split\_1 to Split\_3, FedAvg suffers a substantial 23.88\% drop, further demonstrating the superior robustness of \methodname{}. 

\subsubsection{Performances on the Face Dataset}

To further demonstrate the efficiency of \methodname{} in more complex scenarios, we compare the performance of different algorithms on a healthcare Face dataset with the multi-label disease task, where AUC is a more common indicator to diagnose positive cases in the medical field. Our experiments in Table \ref{tab:Face} illustrate that \methodname{} (MAS) greatly reduces the average communication cost by 32.06\% and improves AUC by 2.18\% compared to the best-performing baseline FedPEFT-LoRA. In particular, \methodname{} outperforms other methods for the diagnosis of Gout and Malignancy Colorectum with 0.7962 and 0.8288 AUC, respectively. Due to the large-scale pre-trained Query2Label model, FedGLF and FedBug have a slight improvement in communication efficiency. Similarly, the sparse fine-tuning of SLoRA in stage 1 seriously affects the efficiency. Although FedPEFT-LoRA achieves superior performance, \methodname{} achieves the most advantageous trade-off between efficiency and effectiveness. FedPEFT-LoRA-FA fine-tunes and shares only the projection-up weight $B$, enhancing efficiency at the cost of performance.

\begin{figure*}[t]
    \centering
    \includegraphics[scale=0.5]{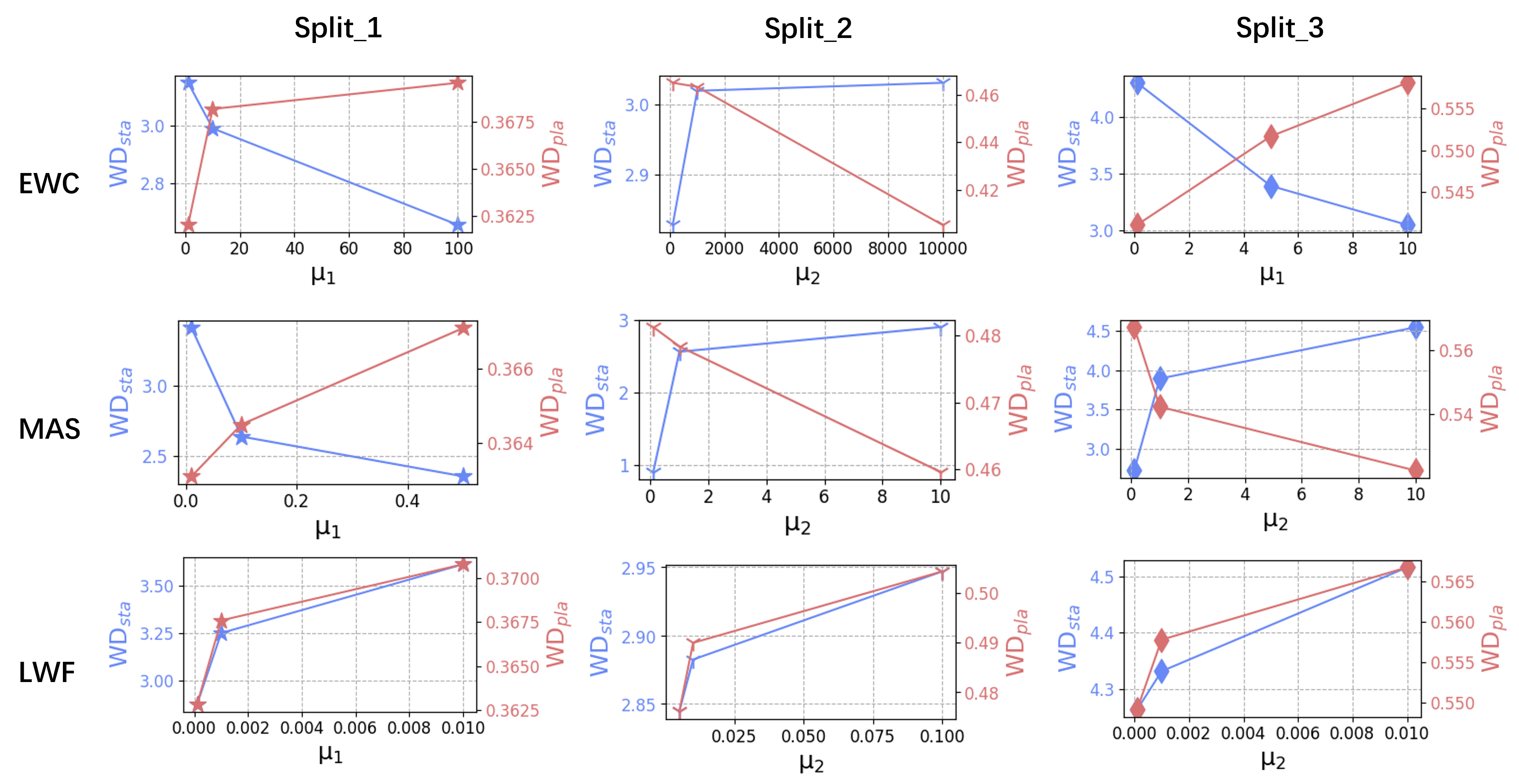}
    \caption{Stability-Plasticity Trade-off Analysis of Weight Distance (WD) with three CL methods under different data partitions.}
    \label{fig:WD}
\end{figure*}

\begin{figure*}[t]
    \centering
    \includegraphics[scale=0.5]{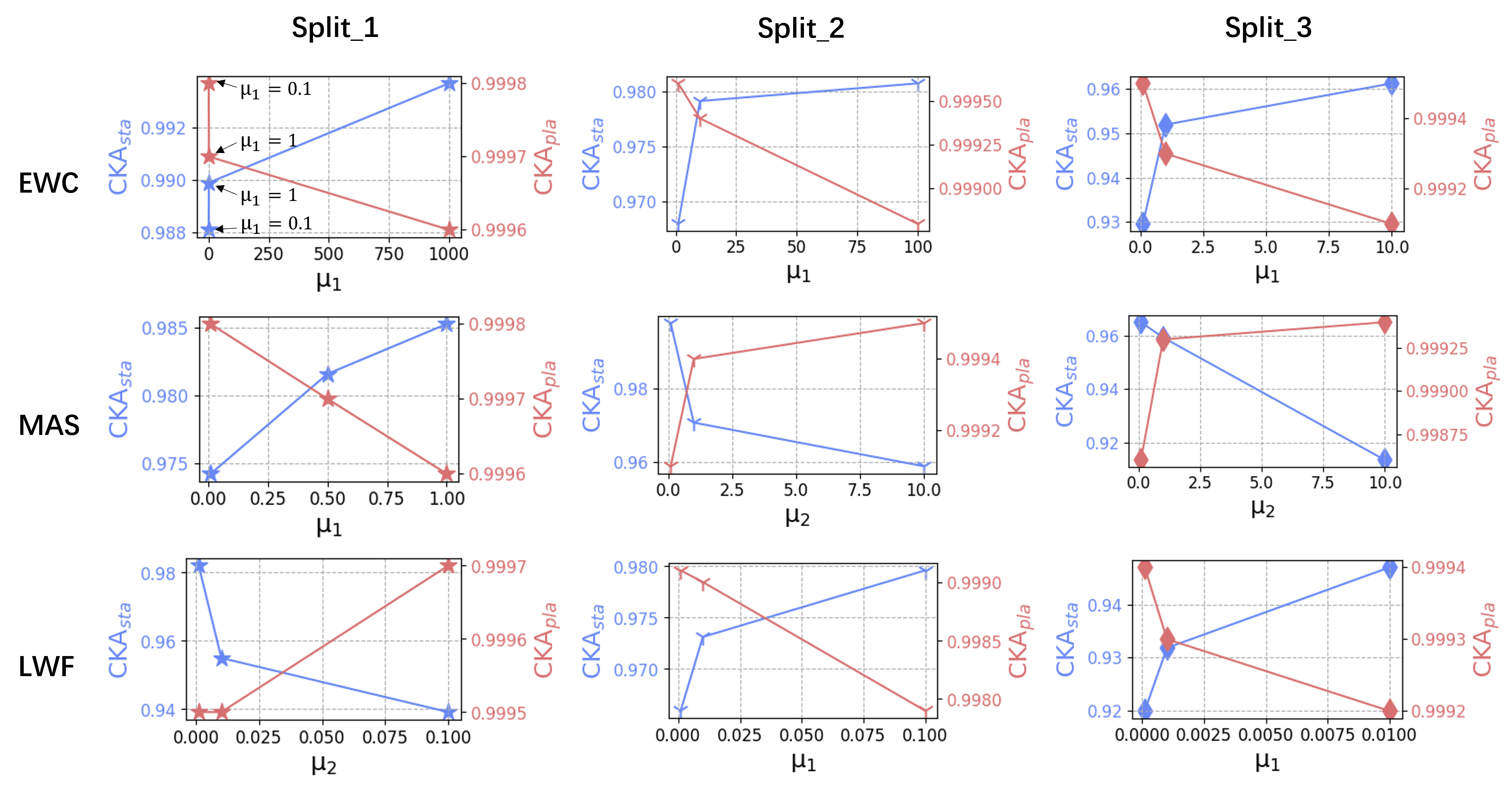}
    \caption{Stability-Plasticity Trade-off Analysis of Centered Kernel Alignment (CKA) with three CL methods under different data partitions.}
    \label{fig:CKA}
\end{figure*}

\subsubsection{Ablation Study}

In this subsection, we conduct an ablation study on the CIFAR-10 dataset to discuss the impact of stability and plasticity regularizers under three different CL methods. Based on gradient analysis, EWC calculates the regularization term by approximating the FIM after each training round. A greater gradient disparity indicates a stronger correlation between the two model parameters. Compared to EWC, MAS defines the regularizer based on the magnitude of parameter updates, prioritizing parameters with minimal changes that induce significant disturbances. LwF is a kind of knowledge distillation method that retains previous knowledge by bridging the gap between the logits of different models. As illustrated in Table \ref{tab:abl_CL}, \methodname{} without CL performs the worst, as inconsistent optimization objectives resulting from increasing parameter dropout rates conflict with each other, thereby degrading model performance. The existence of stability regularizer effectively alleviates the catastrophic forgetting problem, for example, \methodname{} (EWC) only with stability regularizer gives a slight boost in accuracy by about 0.5\% compared to \methodname{} without CL in the Split\_3 setting. The plasticity is less effective than the stability term in most cases because fewer model parameters in the stepwise parameter dropout mechanism determine the new optimization objective. The trade-off of stability and plasticity enhances the compatibility when the two regularizers are mutually suitable.

\subsubsection{Analysis of Weight Distance}

We analyze the stability-plasticity trade-off via WD -- a metric with stronger weight similarity indicates less knowledge forgetting. Figure \ref{fig:WD} illustrates WD values across three CL methods (rows) and different data partitions (columns) as the regularization strength $\mu_1$ or $\mu_2$ is adjusted. Results show that the magnitude of the stability value $WD_{sta}$ (left axis) exceeds that of plasticity $WD_{pla}$ (right axis). Thus, $WD_{sta}$ is always higher than $WD_{pla}$, which implies that the local model has more plasticity and aligns well with the current global model at convergence. EWC and MAS directly regularize the weights, guiding them toward the contrasted weights. As a result, $WD_{sta}$ decreases while $WD_{pla}$ increases with rising $\mu_1$ or decreasing $\mu_2$. On the contrary, the flexible distillation approach LwF learns knowledge based on the logits, resulting in a relatively close to the contrasted weights. Therefore, $WD_{sta}$ and $WD_{pla}$ both increases as $\mu_1$ or $\mu_2$ becomes larger in \methodname{} (LwF).

\subsubsection{Analysis of Centered Kernel Alignment}

Similar to the analysis of WD, we further utilize another metric CKA to analyze the stability-plasticity trade-off, which evaluates the similarity through the representation instead of the model weights. We can see from Figure \ref{fig:CKA} that increasing $\mu_1$ leads to superior stability (higher $CKA_{sta}$) and inferior plasticity (lower $CKA_{pla}$). Naturally, as $\mu_2$ increases, the representations of the local model diverge from those of the accumulated previous global model (lower $CKA_{sta}$) while converging toward the current global model (higher $CKA_{pla}$). In addition, the change range of $CKA_{sta}$ is rather big compared to that of $CKA_{pla}$, indicating the plasticity is steadier when the model achieves convergence.

\section{Conclusion and Future Work}

In this article, we focus on the communication efficiency of FL, where existing works optimize the model performance guided by a given parameter dropout rate in advance. To seek a suitable dropout rate with the prescriptive performance, we propose the \methodname{} approach. It performs stepwise parameter dropout on the server side by leveraging SGC as a signal for adaptive parameter dropout. To mitigate performance degradation caused by inconsistent optimization objectives from varying dropout rates, \methodname{} incorporates continual learning on the client side to balance stability and plasticity. Experimental evaluation on the natural CIFAR-10 dataset and a suite of real-life Face datasets demonstrates its advantages over six state-of-the-art existing methods in terms of performance enhancement and communication cost reduction.

In future research, we plan to explore combining \methodname{} with other PEFT methods than just LoRA (e.g., Adapters), as well as deploying it with popular large language models and vision-language models. Additionally, developing a parameter restoration mechanism to prevent performance collapse caused by excessive dropout is a promising direction worth investigating.

\section*{Ethical Statement}
Institutional Review Board (IRB) / Ethics Committee approvals were obtained in all locations and all participating subjects of the Face dataset have signed a consent form.

\bibliographystyle{IEEEtran}
\bibliography{ref}

\end{document}